\documentclass{article} 
\usepackage[dvipsnames]{xcolor}
\usepackage{iclr2023_conference,times}
\pdfoutput=1

\usepackage{amsmath,amsfonts,bm}

\def\eqref#1{equation~\ref{#1}}

\def\1{\bm{1}}

\DeclareMathAlphabet{\mathsfit}{\encodingdefault}{\sfdefault}{m}{sl}
\SetMathAlphabet{\mathsfit}{bold}{\encodingdefault}{\sfdefault}{bx}{n}

\definecolor{citecolor}{HTML}{0071BC}
\definecolor{linkcolor}{HTML}{ED1C24}
\definecolor{alizarin}{rgb}{0.82, 0.1, 0.26}
\usepackage[pagebackref=false, breaklinks=true, letterpaper=true, colorlinks, citecolor=citecolor, linkcolor=linkcolor, bookmarks=false]{hyperref}
\usepackage{url}
\usepackage{graphicx}
\usepackage{multirow}
\usepackage{booktabs}
\usepackage{makecell}

\usepackage{paralist}
\usepackage{threeparttable}
\usepackage{enumitem}
\usepackage{makecell}
\usepackage{wrapfig}
\usepackage{tabu}
\usepackage{bm}
\usepackage{caption}

\usepackage{color, colortbl}

\def\Modelname{\textsc{DaM}}
\title{\Modelname: Dynamic Adapter Merging for Continual Video QA Learning}

\def\aspace{\hspace{0.4em}}

\author{\vspace{1mm} Feng Cheng$^\dag$ \aspace Ziyang Wang$^\dag$ \aspace Yi-Lin Sung \aspace Yan-Bo Lin \aspace Mohit Bansal \aspace Gedas Bertasius \\
\vspace{1mm} Department of Computer Science, UNC Chapel Hill \\
{\vspace{1mm} \tt\small \{fengchan,ziyangw,ylsung,yblin,mbansal,gedas\}@cs.unc.edu} \\
{\vspace{1mm} \small $^\dag$ Equal Contribution}
}

\iclrfinalcopy 
\begin{document}

\maketitle

\begin{abstract}
We present a parameter-efficient method for continual video question-answering (VidQA) learning. Our method, named \Modelname, uses the proposed \textbf{D}ynamic \textbf{A}dapter \textbf{M}erging to (i) mitigate catastrophic forgetting, (ii) enable efficient adaptation to continually arriving datasets, (iii) handle inputs from unknown datasets during inference, and (iv) enable knowledge sharing across similar dataset domains. Given a set of continually streaming VidQA datasets, we sequentially train dataset-specific adapters for each dataset while freezing the parameters of a large pretrained video-language backbone. During inference, given a video-question sample from an unknown domain, our method first uses the proposed non-parametric router function to compute a probability for each adapter, reflecting how relevant that adapter is to the current video-question input instance. Subsequently, the proposed dynamic adapter merging scheme aggregates all the adapter weights into a new adapter instance tailored for that particular test sample to compute the final VidQA prediction, mitigating the impact of inaccurate router predictions and facilitating knowledge sharing across domains. Our \Modelname~model outperforms prior state-of-the-art continual learning approaches by 9.1\% while exhibiting 1.9\% less forgetting on 6 VidQA datasets spanning various domains. We further extend \Modelname~to continual image classification and image QA and outperform prior methods by a large margin. The code is publicly available at: \url{https://github.com/klauscc/DAM}.
\end{abstract}

\section{Introduction}
The role of video in our lives has increased tremendously over the recent years, with millions of hours of video uploaded to the Internet daily. Due to such rapid video growth and the emergence of video-language models~\citep{yu2021learning,yang2022zero,cheng2023vindlu, Wang_2023_ICCV,pramanick2023egovlpv2,pramanick2023jack}, video question-answering (VidQA) has become one of the most important tasks in video understanding. However, modern VidQA models often assume static conditions with fixed training datasets. In contrast, many real-world applications increasingly demand adaptability to distribution shifts of continually arriving datasets. For instance, a VidQA model trained only on movie videos may struggle when questioned about instructional or social media videos due to stark domain disparities. Additionally, even within a single domain, a model trained on videos collected before 2020 may fail to answer questions about videos recorded in 2024 due to a substantial time difference between training and testing videos. 

One could address these issues by fine-tuning a VidQA model each time new data is introduced. However, it would cause the model to forget previously acquired knowledge -- a phenomenon commonly referred to as \textit{catastrophic forgetting}~\citep{mcclelland1995there,mccloskey1989catastrophic}. An alternative strategy is to retrain the model by incorporating both existing training data and the newly acquired data. However, training the model on the combined data is impractical due to the even larger computational cost~\citep {zellers2021merlot,fu2021violet,li2023lavender,wang2022omnivl}. These challenges underscore the necessity for \textit{continual VidQA learning}, where the VidQA model gradually learns to incorporate knowledge from continuously evolving video training data with minimal training cost.

\begin{figure}[t]
    \centering
    \includegraphics[width=\textwidth]{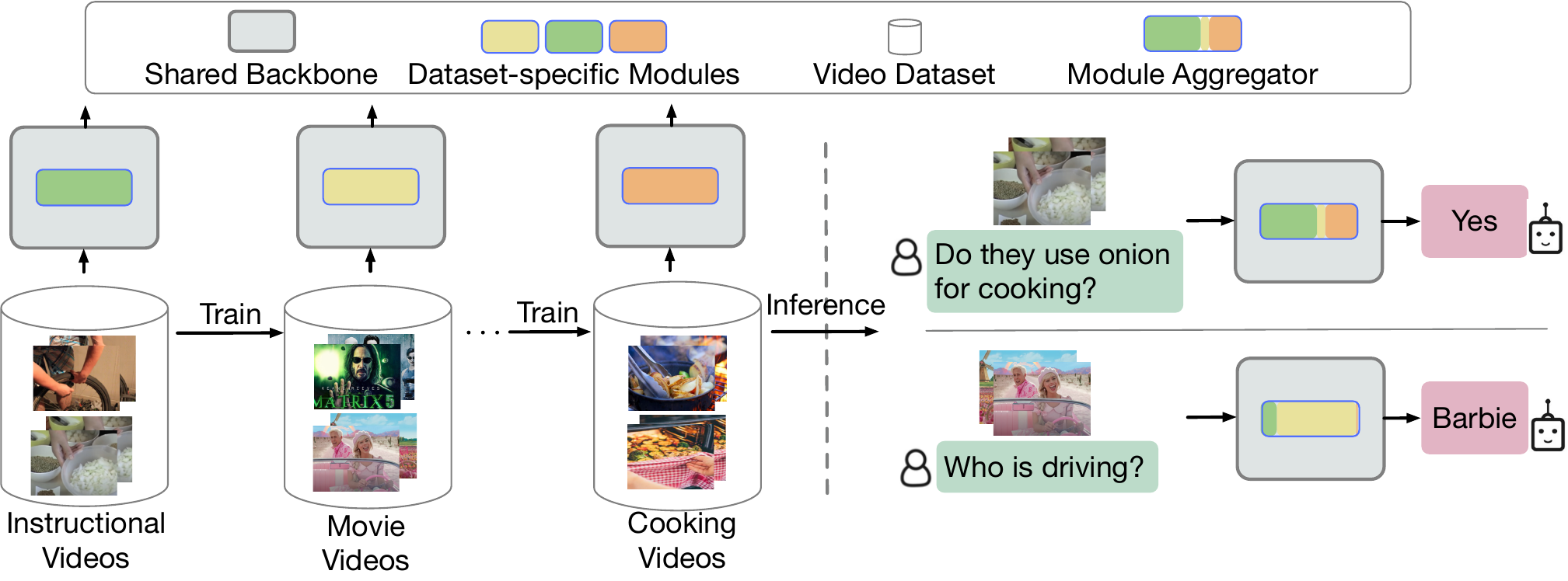}
    \caption{A high-level overview of our proposed Domain-Incremental Learning (DIL) framework for Video Questions-Answering (VidQA). Our model is continually trained on sequentially arriving datasets and evaluated on test samples with unknown dataset identities. Our framework \textbf{(i)} incorporates dataset-specific modules to allow specialization and mitigate forgetting, \textbf{(ii)} enables efficient adaptation to continually streaming datasets, \textbf{(iii)} ensures robustness to incorrect module selections, and \textbf{(iv)} facilitates knowledge-sharing across similar datasets.}
    \label{fig:teaser}
\end{figure}

In this work, we focus on the Domain-Incremental Learning (DIL) subproblem of continual learning~\citep{kirkpatrick2017overcoming,wang2023comprehensive}, since it matches the above-discussed challenges of continuously adapting to datasets spanning different domains and time shifts. The key challenges in DIL arise from distribution shifts between sequentially-arriving training datasets. When the distribution shifts between datasets are large, the optimal representation for each distribution can be very different, thus leading to poor performance among regularization-based DIL methods~\citep{kirkpatrick2017overcoming,li2017learning}, which use fully shared parameters across datasets. Recent prompt-based approaches~\citep{wang2022s,wang2022learning} alleviate this issue by using dataset-specific prompts independently trained on each dataset. During inference, these methods rely on a router function to predict the dataset identity and select the corresponding prompts. However, when distribution shifts between datasets are subtle, predicting dataset identity becomes challenging, adversely affecting the performance of such methods. Additionally, selecting individual dataset-specific prompts prevents knowledge-sharing between datasets, which may be suboptimal when the training datasets are similar. Thus, as shown in Fig.~\ref{fig:teaser}, an ideal DIL method should \textbf{(i)} incorporate dataset-specific modules to allow specialization and limit catastrophic forgetting, \textbf{(ii)} enable efficient adaptation to continually arriving datasets, \textbf{(iii)} be robust to incorrect dataset-specific module selections, and \textbf{(iv)} facilitate knowledge-sharing across similar domains.

Motivated by these observations, we propose \textbf{D}ynamic \textbf{A}dapter \textbf{M}erging (\Modelname), a highly-performant, generalizable, and parameter-efficient continual VidQA learning scheme. Our model consists of (i) an adapter for each continually arriving dataset, (ii) a non-parametric router, and (iii) a dynamic adapter merging module.
Given a sequence of VidQA datasets spanning different data distributions, we begin by training a \textit{dataset-specific adapter} for each dataset while freezing the pretrained video-language backbone (e.g., CLIP~\citep{radford2021learning} and DeBERTa~\citep{he2020deberta}).
Afterward, given a test sample from an unknown dataset during inference, we use a non-parametric video-language router to estimate probabilities for each dataset-specific adapter. These probabilities reflect the relevance of each adapter to that particular video-question input instance. Subsequently, the proposed dynamic adapter merging module merges all the adapter weights into a new adapter instance tailored for that particular test sample to compute the final VidQA prediction. As a result, even if the router produces partially inaccurate probabilities, \Modelname~could still answer the VidQA problem as our dynamic merging scheme incorporates knowledge from multiple adapters, often including those associated with the correct domain. Therefore, the proposed dynamic merging scheme mitigates the impact of inaccurate router predictions and also facilitates knowledge sharing across distributions, thereby enhancing VidQA performance.

Our \Modelname~method outperforms prior prompt-based DIL models~\citep{wang2022s,wang2022learning} by \textbf{9.1\%} on 6 sequentially-introduced VidQA datasets from various domains while exhibiting \textbf{1.9\%} less forgetting. \Modelname~can also be easily extended to tasks such as image classification ( +\textbf{9.32\%} on CORe50) and image question-answering (+\textbf{4.4\%} on a benchmark with 4 datasets). Furthermore, we conduct extensive ablation studies to analyze the relationship between dynamic merging and router, elucidating the key success factors of our approach. We will release our code and pretrained models to enable the community to develop models for this emerging research area of domain-incremental VidQA learning.

\section{Related Work}

\textbf{Video Question Answering (VidQA)} represents a fundamental task in video-language understanding, aiming to answer natural language questions from video inputs. Most commonly used methods \citep{yang2022zero, yu2023self,xiao2022video, cheng2023vindlu,lei2021less,li2020hero,miech2019howto100m,sun2019videobert} leverage video-language models (VLMs) with transformer architecture \citep{xiao2022video, lei2021less, cheng2023vindlu} and large pre-trained language models \citep{yang2022zero, yu2023self}. FrozenBiLM \citep{yang2022zero} handles the multimodal input using a pretrained bidirectional language model and casts VidQA as a masked language modeling problem. SeViLA \citep{yu2023self} builds upon a large image-language model, BLIP-2 \citep{li2023blip2}, and extends it to accommodate video input for VidQA.  However, none of these methods are designed to handle continual shifts in training data distribution, which is our focus in this work.

\noindent \textbf{Continual Learning (CL)} focuses on developing frameworks that can continually learn from streaming training datasets. This is a fundamental challenge for many deep learning methods due to \textit{catastrophic forgetting}~\citep{mcclelland1995there}. Continual learning methods can be categorized into regularization-based approaches~\citep{kirkpatrick2017overcoming,li2017learning}, replay-based approaches~\citep{cha2021co2l,riemer2018learning}, optimization-based approaches~\citep{lopez2017gradient,chaudhry2018efficient} and representation-based approaches~\citep{gao2023unified,foret2020sharpness,ermis2022memory,douillard2022dytox}.
Several recent CL approaches use pre-trained models for the vision-language domain, including CLiMB~\citep{srinivasan2022climb} for task-incremental learning, VQACL~\citep{zhang2023vqacl} and CL-CrossVQA~\citep{zhang2022cl} for rehearsal-based Domain-Incremental Learning (DIL).
Rehearsal-based methods require storing data of previously used training datasets, which may not be possible in real-world settings due to privacy or intellectual property concerns. In contrast, rehearsal-free CL approaches \citep{li2017learning,smith2023closer,smith2021always, Jung_2023_ICCV, li2023learning, zuo2024hierarchical, wang2023orthogonal, wang2023general} do not require storing any previous training data. Among these, several recent prompt-based methods such as L2P~\citep{wang2022learning}, DualPrompt~\citep{wang2022dualprompt}, S-Prompts~\citep{wang2022s} and CODA-Prompt~\citep{smith2023coda} used visual prompts~\citep{liu2023pre} prepended to a pre-trained transformer and extended prompt-based learning for continual learning scenarios. Unlike these prior prompt-based DIL methods, we propose dynamic adapter merging to alleviate the issues of inaccurate router predictions and enable cross-domain knowledge sharing.

\noindent \textbf{Model Merging} aims to merge multiple domain models into a single model that can be used for inference on these domains. The work in~\citep{wortsman2022robust,ilharco2022patching} computes the merged weights as an element-wise arithmetic mean of the weights of all domain models. Subsequently, several methods proposed to improve the performance of the model merging using techniques such as Fisher Merging~\citep{matena2022merging}, RegMean~\citep{jin2022dataless}, Git Re-Basin~\citep{ainsworth2022git}, Task Arithmetic~\citep{ilharco2022editing} and TIES-Merging~\citep{yadav2023resolving}.
Model merging has been applied to many scenarios, including federated learning~\citep{mcmahan2017communication}, improving out-of-domain generalization~\citep{cha2021swad}, and improving performance on a single target task~\citep{gupta2020stochastic,wortsman2022model}. 
Recently, the method in~\citep{GuerreroPea2022RebasinVI} proposed a Sinkhorn re-basin network for replay-based class incremental continual learning but only experimented with small models (e.g., ResNet18~\citep{he2016deep}) on small datasets (e.g., CIFAR-100~\citep{krizhevsky2009learning}).
Unlike existing model merging methods that create a single merged model for all datasets, we dynamically generate a new model instance tailored for each test sample with minimal computational overhead.
\section{Technical Approach}

\subsection{Unified Formulation}
\label{sec:unified_formulation}

We first consolidate recent DIL methods~\citep{wang2022s,wang2022learning} into a unified formulation. Specifically, most existing DIL methods share a common structure comprising a frozen pretrained backbone $f_\theta$ with parameters $\theta$, dataset-specific modules (i.e., prompts) $M = \{m_1,...,m_T\}$, and a router. For the dataset arriving at time $t$, only the dataset-specific module $m_t$ is trained, while the backbone $f_\theta$ and all previously learned modules $m_1,...,m_{t-1}$ are frozen to prevent forgetting. During inference, given a test sample $x$ with an unknown dataset identity, the inference process is formulated as Eqn.~\ref{eqn:unified_formulation}, where $f_{\theta, m_i}$ represents the pretrained backbone augmented with a dataset-specific module $m_i$, $p$ is the predicted probability depicting how relevant $x$ to each dataset-specific module,  and $x$ and $y$ denote the input and output, respectively. 
\begin{equation}
  \label{eqn:unified_formulation}
\begin{aligned}[c]
  p &= \text{router}(f_\theta(x)) \\
  i &= \text{argmax}(p) \\
  y &= f_{(\theta, m_{i})}(x)
\end{aligned}
\end{equation}
We identify suboptimal aspects in the formulation of Eqn.~\ref{eqn:unified_formulation}, notably (i) potential errors introduced by the router's incorrect probability predictions leading to the erroneous selection of a dataset-specific module $m_i$ and (ii) the lack of knowledge-sharing among modules $m_1,...,m_T$.

Next, in Eqn.~\ref{eqn:unified_formulation2}, we propose a more general formulation that replaces the argmax operation with a composer function. Rather than selecting a single module corresponding to the highest router probability, the composer aggregates multiple dataset-specific modules into one, offering robustness to imperfect router predictions and enabling knowledge sharing among dataset-specific modules. 
\begin{equation}
  \label{eqn:unified_formulation2}
\begin{aligned}[c]
  p &= \text{router}(f_\theta(x)) \\
\bm{m'} &= \bm{\textbf{composer}(M, p)} \\
  y &= f_{(\theta, m')}(x)
\end{aligned}
\end{equation}

In the next section, we describe how we instantiate our above-described general formulation with specific modeling components. 

\begin{figure}[tbp]
    \centering
    \includegraphics[width=\textwidth]{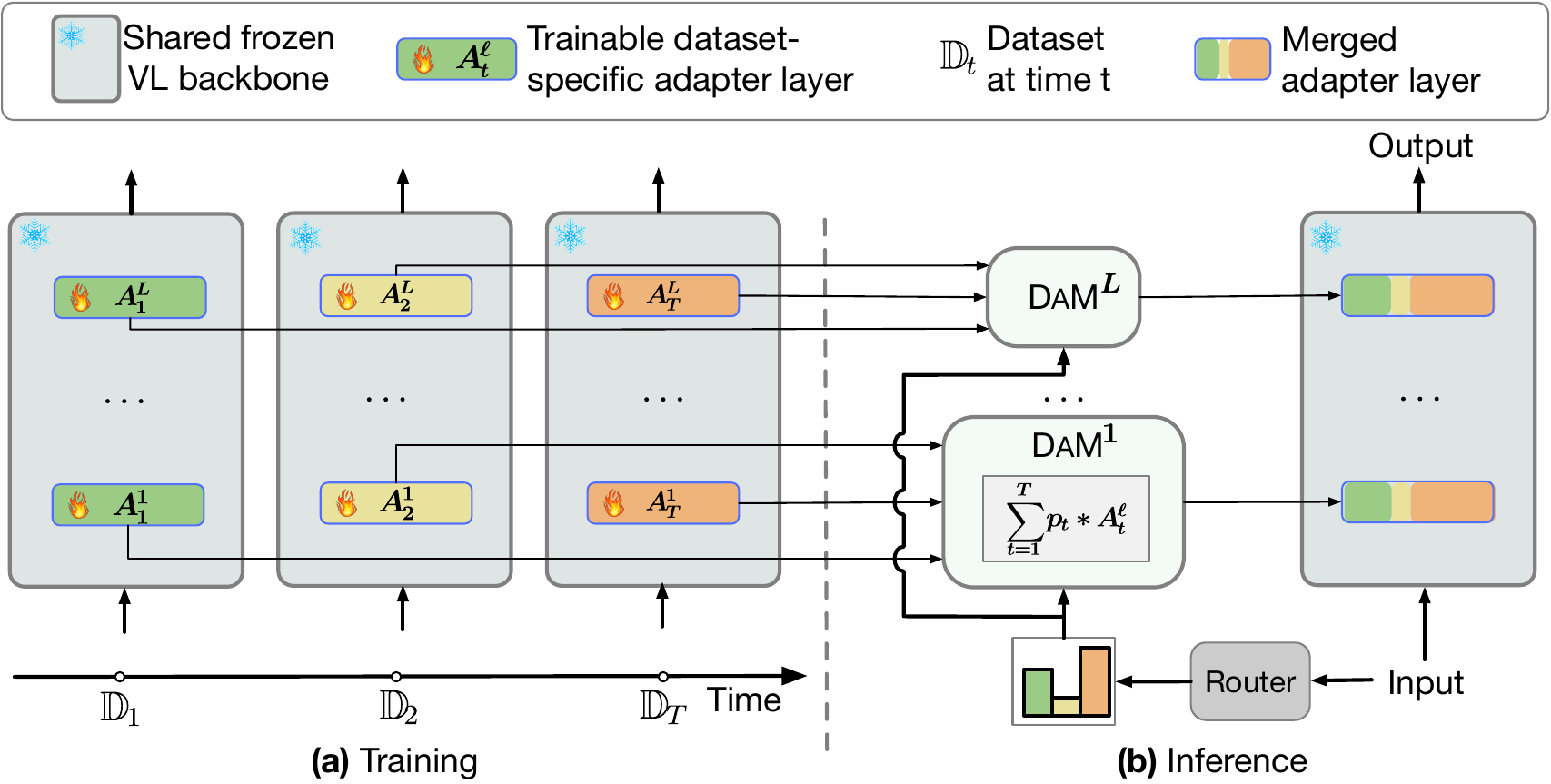}
    \caption{An overview of our Dynamic Adapter Merging (\Modelname) framework. \textbf{(a)} Our model is continually trained on sequentially arriving datasets $\{\mathbb{D}_1,..., \mathbb{D}_T\}$. During training on dataset $\mathbb{D}_t$, we only train the adapter $A_t = \{A_t^{(\ell)}\}_{\ell=1}^{L}$ while keeping previously learned adapters fixed.
      \textbf{(b)} During inference, given a test sample (a video and a text question), we use the proposed router to predict the probability of each adapter being relevant to that particular input instance. Afterward, we dynamically merge multiple dataset-specific adapters in parameter space to reduce the impact of incorrect router predictions and leverage cross-domain VidQA cues. Finally, the pretrained backbone, together with the merged adapter, is used to make the final VidQA predictions.}
    \label{fig:method overview}
\end{figure}

\subsection {\Modelname: Dynamic Adapter Merging}

Based on the formulation in Eqn.~\ref{eqn:unified_formulation2}, we propose \Modelname~, a framework that can learn to model sequentially streaming data $\mathbb{D}_t$ with little forgetting and minimal computational overhead. As shown in Fig.~\ref{fig:method overview}, our method consists of four main components: (i) a frozen pretrained video-language backbone $f_\theta$, (ii) dataset-specific modules $m_1,...,m_T$ implemented as adapters~\citep{houlsby2019parameter} for each training dataset, (iii) a non-parametric video-language router that predicts probabilities for selecting the most relevant adapters for a given test-time VidQA input instance, and (iv) a dynamic adapter merging scheme as a composer to aggregate all the adapter modules. We now describe each component in more detail.

\noindent \textbf{Backbone.} Our backbone $f_\theta$ is a large-scale pretrained VidQA model (e.g. FrozenBiLM~\citep{yang2022zero}), implemented with Transformers (e.g. CLIP ViT-L/14~\citep{radford2021learning} and DeBERTa-V2-XL~\citep{he2020deberta}). In practice, our framework can be applied to arbitrary backbones as shown in Sec.~\ref{sec:exp:gen to images}.

\noindent \textbf{Dataset-Specific Modules.} 
\label{sec:method:adapter}
We propose implementing our dataset-specific modules $m_1,...,m_T$ using adapters~\citep{houlsby2019parameter}. These adapter modules are then used to learn from sequentially streaming data $\mathbb{D}_t$, introduced to the model at time $t$. Compared to visual prompts, commonly used in DIL methods~\citep{smith2023coda,wang2022s,wang2022learning}, adapters have several important advantages. First, the larger modeling capacity of adapters is beneficial as it enables us to effectively capture distribution-specific information from each dataset $\mathbb{D}_{t}$. Furthermore, the high parameter efficiency of adapters (e.g., $\sim3\%$ of total parameters in a pretrained backbone) allows us to efficiently train our framework every time new data $\mathbb{D}_{t}$ is received. 

Specifically, we use an adapter $A_t = \{A_t^{(\ell)}\}_{\ell=1}^{L}$ consisting of $L$ adapter layers for each sequentially streaming dataset $\mathbb{D}_{t}$. We use the standard adapter structure as in~\citep{houlsby2019parameter,yang2022zero}, and insert an adapter layer $A_t^{(\ell)}$ after each self-attention and feed-forward network layer in our pretrained backbone. During training on dataset $\mathbb{D}_{t}$, we only train an adapter $A_t$  while keeping previously learned adapters fixed. This allows each adapter to focus on a single dataset, which is advantageous for maximizing dataset-specific performance while alleviating catastrophic forgetting. Additionally, to inherit previously acquired knowledge, we initialize $A_t$ with the weights of adapter $A_{t-1}$ trained in the preceding time step $t-1$, which we denote as \textit{continual initialization}.

\noindent \textbf{Router.}
\label{sec:method:router}
To handle unknown dataset identity during inference, we employ a non-parametric router to predict the probability for each adapter, estimating their relevance to a given video-question input instance from an unknown distribution. Specifically, during training, we calculate the centroid $c_t = \frac{1}{N_t^s}\sum_{i=1}^{N_s} f_\theta(x_{t,i}^s) \in \mathbb{R}^{d}$ of each dataset $\mathbb{D}_t$ by averaging all multimodal video-language features extracted by the frozen pretrained backbone $f_\theta$ without adapters. Then, for a test sample $x$ during inference, we calculate adapter-specific probabilities $p_t = \frac{\text{exp}(l_t / \tau)}{\sum_{i=1}^{T}\text{exp}(l_i / \tau)}$. Here, $l_t = \text{cos}(f_\theta(x), c_t)$ is the cosine similarity between a feature $f_\theta(x)$ and a centroid $c_t$, $T$ is the total number of datasets up to the current point, and $\tau$ is the temperature hyperparameter.
We find our simple non-parametric router is more effective and computationally cheaper than other more complex design choices, including the ones used in prior DIL methods~\citep{smith2023coda,wang2022learning}, as we will show in Sec.~\ref{sec:analysis_router}.

\noindent \textbf{Composer.}
\label{sec:method:merging}
To improve our DIL framework's robustness to incorrect router predictions and enable knowledge-sharing across dataset-specific modules, we implement our composer function drawing the ideas from the model merging research community~\citep{jin2022dataless,ainsworth2022git,yadav2023resolving}. In particular, recent model merging techniques~\citep{wortsman2022model,jin2022dataless} have demonstrated the effectiveness of merging multiple domain models in the parameter space into a single model that generalizes to all the merged domains, thus, effectively eliminating the need for dataset identity predictions and naturally leveraging knowledge-sharing. However, a single fixed model may lack the expressiveness required to handle numerous domains, as observed in~\citep{yadav2023resolving}, where the performance of the merged model drops significantly (e.g., $>10\%$) when the number of domains is large (e.g., 8 domains).

Motivated by these considerations, we implement our composer using our proposed Dynamic Adapter Merging (\Modelname) scheme that dynamically merge multiple dataset-specific adapters for each test-time input instance (Figure~\ref{fig:method overview}b). Our composer is implemented through a simple instance-wise adapter weight merging using soft router-predicted probabilities. Note that all dataset-specific adapters share the same architecture, enabling element-wise merging of all adapters in their parameter space. Specifically, given adapter weights for all $T$ observed datasets $\mathcal{A} = \{A_1, \hdots, A_T \}$, and input-specific router probabilities $p \in \mathbb{R}^{T}$, the merged adapter weights $A_{M} = \sum_{t=1}^{T}p_t \cdot A_{t}$ are incorporated with the pretrained backbone for the final VidQA prediction.

Our dynamic adapter merging scheme is advantageous since it enhances robustness to incorrect dataset identity predictions. In particular, even when the router function produces partially incorrect dataset-identity probability predictions for the adapter selection, our dynamic merging scheme incorporates knowledge from multiple adapters, often including those associated with the correct domain. Additionally, such a dynamic adapter merging scheme facilitates knowledge sharing between dataset-specific adapters through parameter-space merging, proving beneficial when multiple datasets stem from similar domains. Unlike existing model merging techniques~\citep{wortsman2022model,jin2022dataless}, which produce a single fixed model for all test samples, our method produces a model that is uniquely tailored for each test sample, thus, offering greater modeling expressivity.

\section{Experiments}

\subsection{Experimental Setup}
\label{sec:experimental setup}

\textbf{Datasets and Metrics.} We perform experiments on 9 Video Question Answering (VidQA) datasets, which include iVQA~\citep{yang2021just}, MSVD-QA~\citep{xu2017video}, MSRVTT-QA~\citep{xu2017video}, LSMDC~\citep{maharaj2017dataset}, ActivityNet-QA~\citep{yu2019activitynet}, TGIF-QA~\citep{jang2017tgif}, TrafficQA~\citep{xu2021sutd}, EnvQA~\citep{gao2021env} and AGQA~\citep{grunde2021agqa}. MSVD-QA, MSRVTT-QA, and ActivityNet-QA involve social media videos, with ActivityNet-QA featuring notably longer videos (i.e., on average 2 minutes in length versus 30 second average duration of the videos in the first two datasets). iVQA, LSMDC, TGIF-QA, TrafficQA, EnvQA, and AGQA represent instructional, movie, short-GIF, traffic, virtual, and indoor human videos, respectively. We train the models on these sequentially arriving datasets. After training the model on the last dataset, we evaluate the resulting checkpoint on the test set of all trained datasets. During the evaluation, the dataset identity of each testing sample is assumed to be unknown. Following \citep{wang2022continual,wang2022s}, we use the \textbf{average accuracy} and \textbf{forgetting} as the evaluation metrics. See Appendix \ref{sup:eval_metrics} for formal definitions.

\noindent \textbf{DIL Baselines.} For all of our continual learning baselines (including our approach), we use the recent FrozenBiLM~\citep{yang2022zero} VidQA model, implemented using CLIP ViT-L/14~\citep{radford2021learning} and DeBERTa-V2-XL~\citep{he2020deberta} video and language backbones and containing 1.2B parameters in total. In our comparisons, we include three recent Prompt-based methods \textit{L2P}~\citep{wang2022learning}, \textit{CODA-Prompt}~\citep{smith2023coda}, and \textit{S-Prompts}~\citep{wang2022s}, and two regularization-based methods, \textit{EwC}~\citep{kirkpatrick2017overcoming} and \textit{LwF}~\citep{li2017learning}. We also incorporate several naive baselines: (i) \textit{Zero-Shot}, which directly evaluates the pretrained model on all datasets without any training, and (ii) \textit{Seq-FT}, which sequentially finetunes the model on the sequentially arriving datasets. The upper bound performance for the adapter and prompt-based variants is obtained by multitask finetuning jointly on all training datasets.

\newcommand{\red}[1]{\textcolor{Red}{#1}}
\newcommand{\fg}[1]{\scriptsize (\textcolor{Red}{#1})}
\newcommand{\gfg}[1]{\scriptsize (\textcolor{Green}{#1})}
\begin{table}[t]
    \centering
    \caption{Comparison with state-of-the-art on Domain-Incremental VidQA Learning. We obtain the upper-bound performance by multitask finetuning jointly on all the datasets. The zero-shot baseline directly evaluates the pretrained model on all datasets without any training, while the Seq-FT baseline sequentially finetunes a single model on all the datasets. We also reimplement prior methods (EwC, LwF, L2P, CODA-Prompt, S-Prompts) using our strong video-language backbone, as these methods were not initially designed for VidQA. All continual learning methods are trained sequentially from left to right in the table. Our proposed \Modelname~outperforms the current state-of-the-art by 9.1\% while exhibiting 1.9\% less forgetting.
    }
    \label{tab:sota:dil}
    \begin{tabular}{lccccccc}
        \toprule
        \multicolumn{8}{l}{Training Order: iVQA $\to$ MSVD $\to$ MSR-VTT $\to$ LSMDC $\to$ ANet $\to$ TGIF} \\
        \midrule
        \multirow{2}{*}{Method} & \multicolumn{7}{c}{Downstream VidQA Accuracy  (\red{Forgetting}) (\%)} \\
        \cmidrule(lr){2-8}
                                & iVQA & MSVD & MSR-VTT & LSMDC & ANet & TGIF & Avg. \\
        \midrule
        Zero-Shot & 26.8 & 33.0 & 15.0 & 51.5 & 25.5 & 41.9 & \multicolumn{1}{c}{32.3} \\
        Seq-FT & 28.4 & 36.0 & 23.7 & 52.1 & 31.2 & 67.6 & 39.8 \\
        \midrule
        \multicolumn{5}{l}{\textit{Multitask Finetuned (Upper-Bounds)}} \\
        Adapters & 39.7 &	56.6 &	46.7 &	62.9&	42.2&	67.8 &	52.6\\
        Prompt Tuning & 35.0 & 49.0 & 37.1 & 57.4 & 33.9 & 59.2 & 45.3 \\
        \midrule
        \multicolumn{5}{l}{\textit{Regularization-based methods}} \\
        EwC & 29.9 & 39.3 & 25.5 & 54.9 & 32.4 & 67.5 & 41.6 \fg{-10.9} \\
        LwF & 28.3 & 38.2 & 25.8 & 56.4 & 33.6 & \textbf{68.5} & 41.8 \fg{-10.7} \\
        \midrule 
        \multicolumn{5}{l}{\textit{Prompt-based methods}} \\
        L2P & 32.8 & 43.3 & 32.1 & 54.8 & 27.2 & 54.4 &40.8 \fg{-4.6} \\
        CODA-Prompt & 32.9 & 44.8 & 28.7 & 50.7 & 23.9 & 54.7 & 39.6 \fg{-5.7} \\
        S-Prompts & 31.8 & 45.5 & 30.2 & 54.9 & 27.9 & 56.1 & 41.1 \fg{-4.2} \\
        \midrule
        \Modelname~(Ours) & \textbf{39.1} & \textbf{53.6} & \textbf{42.2 } & \textbf{63.0} & \textbf{36.3} & 66.8 & \textbf{50.2 \fg{-2.3}} \\
        \bottomrule
    \end{tabular}
\end{table}

\noindent \textbf{Model Merging Baselines.} We also compare with two model merging methods, Average Merging~\citep{wortsman2022model,ilharco2022editing} and RegMean~\citep{jin2022dataless}. In our implementation, both methods merge all the dataset-specific adapters into a single fixed adapter, which is then applied to all the test samples. This is in contrast to our framework, which produces a uniquely tailored adapter module for each test sample.

We refer readers to Appendix~\ref{sup:implementations} for detailed implementations of our framework and all the baseline methods.

\subsection{Comparison with State-of-the-art}
\label{sec:exp:sota}

\textbf{Comparison with Domain-Incremental Learning (DIL) Methods.} Tab.~\ref{tab:sota:dil} compares our method and state-of-the-art DIL approaches. Our findings demonstrate that our proposed \Modelname~scheme outperforms the leading DIL method, S-Prompts, by a substantial margin of \textbf{9.1\%} in average accuracy while also exhibiting \textbf{1.9\%} less forgetting. Among prompt-based methods, L2P, CODA-Prompt, and S-Prompts show reduced forgetting compared to regularization-based methods EwC and LwF. However, these prompt-based methods achieve lower overall accuracy, which can be attributed to their smaller learning capacity. Overall, these results show the effectiveness of our proposed framework.

\noindent \textbf{Comparison with Model Merging Methods.} Next, we compare our method with two model merging methods, Average Merging~\citep{wortsman2022model} and RegMean~\citep{jin2022dataless}. For a fair comparison, all the methods merge the same set of adapters, each individually fine-tuned on each dataset without our continual initialization scheme.  As shown in Tab.~\ref{tab:dam as model merging}, \Modelname~outperforms RegMean by \textbf{6.0\%} and average merging by \textbf{7.5\%} in average accuracy. We hypothesize that the significantly better performance of our model comes from the fact that \Modelname~produces a custom model instance for each input instance. This makes our methods a lot more expressive than the existing model merging methods that use a single merged model instance for all data samples.

\begin{table}[tbp]
    \centering
    \setlength{\tabcolsep}{4pt}
    \caption{Comparison with existing model merging techniques. For a fair comparison, all the methods merge the same set of adapters, each individually fine-tuned on each dataset without our continual initialization scheme (i.e., using random initialization). Our \Modelname~outperforms existing model merging methods by a large margin on average.}
    \label{tab:compare_with_model_merging}
    \begin{tabular}{lcccccccc}
    \toprule
    Method	&iVQA&	MSVD&	MSR-VTT&	LSMDC	&ActivityNet	&TGIF&	Avg.\\
    \midrule
    Multitask (upper-bound) &	39.7&	56.6&	46.7&	62.9&	42.2&	67.8&	52.6 \\
    \midrule
    Avg. Merging & \textbf{38.0} & 45.7 & 27.7 & 54.5 & 27.0 & 56.6 & 41.6 \\
    RegMean & 36.6 & 49.7 & 32.5 & 54.0 & 27.7 & 57.8 & 43.1 \\
    \Modelname~(Ours)	& 36.5 &	\textbf{51.6} &	\textbf{39.5}	&\textbf{63.0}&	\textbf{36.5}	&\textbf{67.7}&	\textbf{49.1} \\
    \bottomrule
    \end{tabular}
    \label{tab:dam as model merging}
\end{table}

\noindent \textbf{Computational Complexity Analysis.} In addition to standard accuracy metrics, we also analyze the computational cost of our proposed approach. We note that each dataset-specific adapter in \Modelname contributes merely \textbf{2.5\%} of the pretrained model's parameters (CLIP-L/14 + DeBerTa-V2-XLarge), totaling 30M parameters. In terms of computational cost, merging adapter parameters incurs just \textbf{0.09} GFLOPs (30M *(2k-1), k=2 in our case), notably lower than the \textbf{162} GFLOPs required by CLIP-L/14 for a single image processing. Therefore, these results show that our proposed \Modelname~can efficiently adapt to a reasonably large number of continually arriving datasets.

\subsection{Generalization to Images}
\label{sec:exp:gen to images}

To further showcase the generalizability of our approach, we extend \Modelname~to two image tasks: 1) image classification and 2) image question-answering.

\textbf{Image classification.}  We conduct experiments on two standard DIL benchmarks: CORe50 \citep{lomonaco2017core50} and DomainNet \citep{peng2019moment}. CORe50~\citep{lomonaco2017core50} contains 50 categories across 11 domains. Following prior work, we continually train the model on 8 domains (120K images) and evaluate the trained model on the remaining 3 domains (40K images). DomainNet contains 345 categories across 6 domains. The DIL setup on DomainNet is the same as \cite{wang2022s,fini2022self}. Following standard evaluation protocol, we use ViT-B/16~\citep{dosovitskiy2020image} pretrained on ImageNet as our backbone. As shown in Tab.~\ref{tab:image classification}, \Modelname~surpasses the current state-of-the-art S-Prompts by \textbf{9.32\%} and \textbf{18.61\%} using the same ImageNet-pretrained ViT-B/16 backbone on CORe50 and DomainNet, respectively. These results suggest that our model can also be effectively applied to DIL image classification tasks.

\definecolor{Gray}{gray}{0.5}
\begin{table}[tbp]
    \centering
    \caption{Domain-Incremental Learning on image classification. Our upper-bound is obtained by finetuning a shared adapter jointly on all domains. For a fair comparison, we de-emphasize S-Prompts with CLIP encoder since it is pretrained with much more data than the ImageNet-pretrained ViT-B/16 backbone used by our method.}
    \label{tab:image classification}
    \begin{tabu}{lcccc}
    \toprule
     \multirow{2}{*}{\bf Method}  & \multirow{2}{*}{\bf Backbone} & Buffer size & \multicolumn{2}{c}{\bf Avg. Accuracy (\%)} \\
     \cmidrule(lr){4-5}
    & & & CORe50 & DomainNet \\
    \midrule
    Multitask (upper-bound) & ViT-B/16 & - & 94.59 $\pm$ 0.21 & 71.95 \\
    \midrule
    DyTox & ViT-B/16 & 50/class & 79.21 $\pm$ 0.10  & 62.94 \\
    \midrule
    LwF & ViT-B/16 & \multirow{4}{*}{0/class} & 75.45 $\pm$ 0.40 & 49.19\\
    L2P & ViT-B/16 & & 78.33 $\pm$ 0.06 & 40.15 \\
    S-Prompts & ViT-B/16 & & 83.13 $\pm$ 0.51 & 50.62\\
    \rowfont{\color{Gray}}
    S-Prompts & CLIP ViT-B/16 & & 89.06 $\pm$ 0.86 & 67.78\\
    \midrule
    \Modelname (Ours) & ViT-B/16 & 0/class & \textbf{92.45 $\pm$ 0.25} & \textbf{69.23} \\
    \bottomrule
    \end{tabu}
\end{table}
\begin{table}[tbp]
    \centering
    \caption{We extend our proposed \Modelname~method to continual visual question-answering (VQA) task on image datasets. For these experiments, we use the recent BLIP-2 model~\citep{li2023blip2} as our visual-language backbone. The proposed \Modelname~outperforms the existing state-of-the-art method (S-Prompts) by 4.4\% in average accuracy while exhibiting 1.2\% less forgetting.}
    \label{tab:vqa:blip2}
    \begin{tabular}{lccccc}
    \toprule
    Method  & \makecell{OK-VQA \\ (test)} & \makecell{aOK-VQA \\ (val)} & \makecell{GQA \\ (val)} & \makecell{VQAv2 \\(val)} & Avg. \\
    \midrule
    Zero-Shot & 40.7 & 35.7 & 44.0 & 63.1 & 45.9 \\
    Multitask (upper-bound) & 49.2 & 51.8 & 58.7 & 76.2 & 58.8 \\
    \bottomrule
    S-Prompts & 42.9 \fg{-5.3} & 46.1 \fg{-2.2} & 47.3 \fg{-7.1} & 65.3 \fg{-6.0} & 50.4 \fg{-5.2} \\
    \Modelname & \textbf{45.1 \fg{-4.1}} & \textbf{50.4 \fg{-1.4}} & \textbf{54.1 \fg{-4.6}} & \textbf{69.8 \fg{-6.4}} & \textbf{54.8 \fg{-4.0}} \\
    \bottomrule
    \end{tabular}
\end{table}

{\textbf{Image question-answering.} \label{sec:exp:blip2}}
Next, we also extend our model to the visual question-answering (VQA) task on images. We integrate our proposed \Modelname~and the best performing prompt-based baseline S-Prompts with the state-of-the-art VQA model, BLIP-2~\citep{li2023blip}, which uses CLIP ViT-G/14~\citep{radford2021learning} and FlanT5-XL~\citep{chung2022scaling} as its vision-language backbone and has 4.1B parameters in total. We then continually train both models on 4 mainstream VQA datasets: OK-VQA~\citep{marino2019ok}, aOK-VQA~\citep{schwenk2022okvqa}, GQA~\citep{hudson2019gqa} and VQAv2~\citep{goyal2017making}. The results are shown in Tab.~\ref{tab:vqa:blip2}. Our proposed \Modelname~outperforms S-Prompts by \textbf{4.4\%} with \textbf{1.2\%} less forgetting, thus, demonstrating the generality of our approach.

\section{Analysis}
\label{sec:exp:ablations}

\subsection{Adapter Merging Analysis}
\label{sec:exp:dynamic merging}

\newcommand{\add}[1]{\textcolor{Green}{(#1)}}
\newcommand{\dec}[1]{\textcolor{Red}{(#1)}}
\begin{table}[tbp]
	\centering
	\caption{We investigate the number of dataset-specific adapters to merge for best performance. The Top-$K$ adapters are selected according to the highest router predicted probabilities. The first 4 rows depict the downstream VidQA accuracy, whereas the last row is the router accuracy. We highlight the largest accuracy gap between adapter merging and non-merging variants. Merging adapters is typically useful when the router makes many incorrect predictions.}
	\label{tab:number_model_merging}
	\begin{tabular}{lcccccc}
		\toprule
		Top-$K$       & MSVD                         & MSR-VTT                     & ActivityNet                & iVQA                        & TGIF                       & LSMDC                      \\
        \midrule
		1 (no-merging)& 49.0                         & 40.4                        & 37.4                       & 37.5                        & 66.3                       & 62.9                       \\
        \midrule
		2             & 53.6                         & 42.2                        & \textbf{36.3 \dec{-1.1}} & 39.1                        & 66.8                       & 63.0                       \\
		3             & 54.6                         & \textbf{42.4\add{+2.0}} & 34.0                       & 39.3                        & \textbf{67.0\add{+0.7}} & 63.0                       \\
		6 (merge all) & \textbf{54.9\add{+5.9}}  & 41.9                        & 33.0                       & \textbf{39.6\add{+2.1}} & 66.9                       & \textbf{63.1\add{+0.2}} \\
        \midrule
		Router Acc & 51.0                         & 69.6                        & 76.4                       & 81.6                        & 96.1                       & 100                        \\
		\bottomrule
	\end{tabular}
\end{table}

\begin{wrapfigure}{r}{0.4\textwidth}
    \centering
    \includegraphics[width=1.0\linewidth]{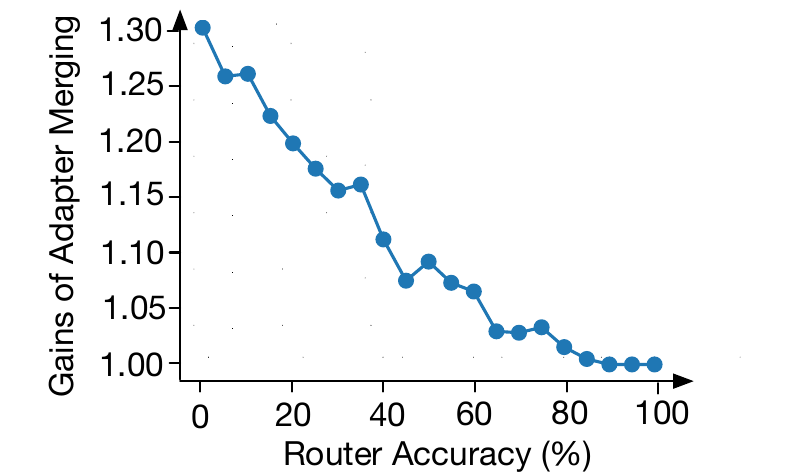}
  \caption{We study the normalized performance gain of dynamic adapter merging as a function of router accuracy. Our results show that dynamic adapter merging leads to a larger boost when the router is inaccurate.}
    \label{fig:model_merging_vs_router}
\end{wrapfigure}
In this section, we analyze the effectiveness of dynamic adapter merging. Specifically, in Tab.~\ref{tab:number_model_merging}, we present a comprehensive breakdown of downstream VidQA accuracy and the router's accuracy on each dataset, considering various adapter merging variants. The table highlights an intriguing trend: as the router's accuracy decreases, the benefits derived from adapter merging become more pronounced.
Specifically, when the router's accuracy is at \textbf{51.0\%} and \textbf{69.6\%}, adapter merging yields substantial downstream accuracy improvements of \textbf{5.9\%} and \textbf{2.0\%} on the MSVD and MSR-VTT datasets, respectively. In contrast, when the router approaches near-perfect accuracy, the gains from adapter merging become less significant (as seen with a marginal \textbf{0.2\%} improvement on LSMDC).

To further validate this observation, Fig.~\ref{fig:model_merging_vs_router} provides insights into the average performance gain of dynamic adapter merging over non-merging variants as a function of router accuracy. The data points are generated by creating a series of routers manually, each predicting dataset probabilities with a specified accuracy. The figure confirms the trend in Tab.~\ref{tab:number_model_merging}, showcasing that adapter merging offers a \textbf{30\%} relative improvement when the router's accuracy drops to \textbf{0\%}.

Based on these results, we can conclude that our proposed \Modelname~is particularly advantageous when dealing with many datasets. In such complex scenarios, dataset prediction becomes notably challenging for the router. These collective findings underscore the practical significance and scalability of our proposed approach in real-world domain-incremental VidQA learning scenarios.

\subsection{Router Analysis}
\label{sec:analysis_router}

In this section, we compare our router design with three router designs from prior DIL methods: L2P~\citep{wang2022learning}, CODA-Prompt~\citep{smith2023coda}, as well as Gaussian Mixture Model (GMM) and Learnable MLP. We incorporate these router functions into our \Modelname~method and measure our model's performance on downstream VidQA task with each of these routers. We also measure the accuracy of each router function for correctly classifying the dataset/domain of a given VidQA input instance. Note that we cannot calculate CODA-Prompts' router's accuracy as it does not explicitly predict the domain identity.
From Tab.~\ref{tab:different_routers}, we observe that higher router accuracy typically leads to higher downstream VidQA accuracy, thus indicating the importance of an accurate router function. Second, we notice that jointly training router and domain-specific modules as was done in previous methods (L2P, CODA-Prompt) leads to worse downstream VidQA accuracy than disjoint training (S-Prompts, Ours). Lastly, our results suggest that despite the simplicity of our non-parametric router function, it produces the best performance.

\begin{table}[tbp]
    \centering
    \setlength{\tabcolsep}{3pt}
    \caption{We study the effectiveness of different router functions. Specifically, we incorporate router functions from several prior methods into our \Modelname~method and measure our model's performance on the downstream VidQA task with each of these routers. Our results suggest that our non-parametric router function leads to the best downstream VidQA performance.}
    \label{tab:different_routers}
    \begin{tabular}{lcccccccc}
    \toprule
    Router	&random&L2P's&CODA-Prompt's&S-Prompts'&GMM&Learnable MLP& Ours \\
    \midrule
    router Acc. & 16.6 & 67.4 & - & 76.4 & 79.0 & 78.7 & \textbf{79.1} \\
    \midrule
    VidQA Acc. & 40.2 & 48.6 & 45.3 & 49.7 & 49.4 & 48.9 & \textbf{50.2} \\
    \bottomrule
    \end{tabular}
    \label{tab: router design choices.}
\end{table}

\definecolor{LightCyan}{rgb}{0.88,1,1}
\definecolor{lemonchiffon}{rgb}{1.0, 0.98, 0.8}
\newcolumntype{t}{>{\columncolor{lemonchiffon}}c}
\newcolumntype{a}{>{\columncolor{LightCyan}}c}

\subsection{Domain Analysis}
\label{sec:exp:domain analysis}
In this section, we analyze the performance of our method through experiments on datasets with both large and small domain gaps. 

\noindent \textbf{Large Domain Gap.} To validate the effectiveness of our framework on datasets with large domain/distribution gaps, we experiment with 4 datasets from 4 different domains: movie videos (LSMDC-QA), indoor human videos (AGQA), traffic videos (TrafficQA), and virtual videos (Env-QA). Tab.~\ref{tab:dil} presents \Modelname’s vidQA accuracy and the router’s domain identity prediction accuracy. We observe that \Modelname~exhibits negligible forgetting on the 4 datasets. We attribute such good performance of our method to 1) dataset-specific adapters that can effectively specialize for modeling each dataset and 2) the high router's accuracy across most datasets in this setting. Consequently, these results indicate that our proposed \Modelname~can be effectively applied to datasets with large domain gaps.

\begin{table}[tbp]
    \centering
    \caption{Domain-Incremental Learning (DIL) on 4 datasets from different domains. \Modelname~has negligible forgetting rate on datasets with large domain gaps.}
    \label{tab:dil}
    \begin{tabular}{lccccc}
    \toprule
    Method  & LSMDC & AGQA &	Env-QA &	TrafficQA($\frac{1}{2}$) & Avg.\\
    \midrule
    Upper-Bound&		63.0&	63.4&	32.3&	67.8&	56.6 \\
    \Modelname	&63.0	&	63.3&	32.0&	67.8&	56.5 \\
    \midrule
    Router Acc. of \Modelname&100	&99.9	&99.2&	99.7&	99.7\\
    \bottomrule
    \end{tabular}
\end{table}
\begin{table}[tbp]
    \centering
    \caption{We evaluate the ability of our framework to adapt to subtle time-distribution shifts. To do this, we divide the iVQA dataset into 5 subsets according to the video upload date to YouTube. We then train our model on these 5 sequentially arriving subsets. Our results indicate that our dynamic adapter merging scheme still works effectively, even when the dataset domains/characteristics are quite similar.}
    \label{tab:dil:distribution-shift}
    \begin{tabular}{lc|c|ccc}
    \toprule
    Method & \makecell{Multitask \\(upper-bound)} & \makecell{Router \\ Acc.} & \makecell{\Modelname \\ (no merging)} & \makecell{\Modelname \\(merging top-$k=2$)} & \makecell{\Modelname \\(merging all)} \\
    \midrule
    VidQA Acc. & 39.8 & 43.8 & 37.1 & 38.4 & \textbf{39.3} \\
    \bottomrule
    \end{tabular}
\end{table}

\noindent \textbf{Small Domain Gap.} Next, we evaluate our approach on datasets within the same domain but collected at different times. Such time-based distribution shifts are typically much smaller than for the previously considered datasets spanning entirely different domains (Tab.~\ref{tab:dil}). Thus, such a setting necessitates knowledge sharing and the ability to handle incorrect router predictions. Specifically, we evaluate our model in this setting by dividing the iVQA dataset into 5 non-overlapping subsets based on the video upload date to YouTube. We continually train the model on these five subsets and then evaluate on iVQA's original test set that spans 5 time distributions. Tab.~\ref{tab:dil:distribution-shift} shows that unlike in the previous setting, in this case, the router attains an accuracy of only 43.8\%. This can be explained by the fact that the dataset/domain-identity prediction problem becomes a lot more challenging due to minor distribution shifts between subsets. In this scenario, the model variant that merges all adapters surpasses the variant without merging by \textbf{2.2\%} and experiences only 0.5\% forgetting. This underscores the effectiveness of dynamic adapter merging and emphasizes the importance of knowledge sharing in settings where domains or datasets are similar.

\subsection{Other Analyses} 

\textbf{Order of the Datasets.} In Tab.~\ref{tab:order_of_domains}, we study how the order of the training datasets affects our model's performance. We randomly sample 5 different orders and train our framework on those orders. Based on the results, we observe that the performance of our approach is quite stable across all 5 orders (\textbf{50.56 $\pm$ 0.26\%}). This indicates our method is robust to the order of training datasets. 

\begin{wraptable}{r}{0.4\textwidth}
    \vspace{-4mm}
    \centering
    \caption{Ablations on the order of datasets. We randomly sampled 5 orders and obtained average accuracies for each order. V: iVQA; D: MSVD; T: MSR-VTT; L: LSMDC; A: ActivityNet; G: TGIF.}
    \begin{tabular}{lccccc}
    \toprule
    Domain Order & Avg. Acc (\%) \\
    \midrule
    V D T L A G & 50.2 \\
    L T G D A V & 50.8 \\
    V A D G T L & 50.4 \\
    G T A V D L & 50.9 \\
    V A G D T L & 50.5 \\
    \bottomrule
    \end{tabular}
    \label{tab:order_of_domains}
\end{wraptable}

\noindent \textbf{Continual Initialization Scheme.} In Section~\ref{sec:method:adapter}, we introduced a continual initialization scheme for initializing a current distribution-specific adapter using the weights of a previously learned adapter. In Tab.~\ref{tab:continual_finetuning}, we validate the effectiveness of this scheme and show that it leads to a notable \textbf{1.1\%} average accuracy improvement. These improvements are particularly pronounced for the datasets that are used first, such as iVQA and MSVD. We posit that the benefits of continual initialization stem from the fact that the weights of continually learned adapters reside in a more similar parameter space. This phenomenon reduces interference disagreements when merging adapters \citep{yadav2023resolving}.

\begin{table}[tbp]
    \centering
    \setlength{\tabcolsep}{4pt}
    \caption{\Modelname~benefits from the proposed continual initialization scheme.}
    \label{tab:continual_finetuning}
    \begin{tabular}{lccccccc}
    \toprule
    Method & iVQA & MSVD & MSR-VTT & LSMDC & ActivityNet & TGIF & Avg. \\
    \midrule
    \Modelname & \textbf{39.1} & \textbf{53.6} & \textbf{42.2} & \textbf{63.0} & 36.3 & 66.8 & \textbf{50.2} \\
    w/o continual initialization & 36.5 & 51.6 & 39.5 & \textbf{63.0} & \textbf{36.5} & \textbf{67.7} & 49.1 \\
    \bottomrule
    \end{tabular}
\end{table}
\section{Discussion and Conclusion}

In this work, we investigate the challenging and relatively unexplored problem of rehearsal-free domain-incremental VidQA learning. Our proposed \Modelname~framework outperforms existing state-of-the-art by \textbf{9.1\%} with $\textbf{1.9\%}$ less forgetting on a benchmark with six distinct video domains. The proposed method \Modelname~is simple and flexible, and we further extend it to image classification tasks and visual question-answering, demonstrating our method's generalization beyond video-level scenarios. Despite effective results, we also observe a few limitations of our proposed approach. Firstly, our approach employs a straightforward weighted averaging technique for merging adapter weights, leaving room for more advanced merging methods that could enhance knowledge sharing among domains. Secondly, our validation encompasses a relatively small number of domains ($\le7$ in our case), consistent with previous domain-incremental learning research. It would be valuable to assess the effectiveness of our method and existing domain-incremental learning methods across a more extensive domain spectrum, potentially involving a substantial number of domains (e.g., 100). We plan to explore these research directions in our future work.

\noindent\textbf{Acknowledgements.} We thank Md Mohaiminul Islam, Ce Zhang, Yue Yang, and Soumitri Chattopadhyay for helpful discussions. This work was supported by the Sony Faculty Innovation award, Laboratory for Analytic Sciences via NC State University, ONR Award N00014-23-1-2356, ARO Award W911NF2110220, DARPA ECOLE Program No. \#HR00112390060, and NSF-AI Engage Institute DRL-2112635. The views contained in this article are those of the authors and not of the funding agency.

\appendix
\section*{Appendix}

In this appendix, we present the following:
\begin{enumerate}[label=\Alph*.]
    \item Implementation details. 
    \item Evaluation Metrics.
    \item Extension to other types of continual learning.
    \item Dataset descriptions.
\end{enumerate}

\section{Implementation Details}
\label{sup:implementations}

\textbf{Details of our \Modelname{} approach.} Our choice for the VidQA model is FrozenBiLM~\citep{yang2022zero}, a state-of-the-art (SOTA) model in the VidQA domain. To align with this model, we utilize a vocabulary encompassing the 3635 most frequent answers. Adhering to the FrozenBiLM approach, we integrate adapters into each layer of the DeBERTa-XL~\citep{he2020deberta} language model, employing a downsampling rate of 8. The loss function is the same as the original FrozenBiLM model, i.e., the cross-entropy loss between the predicted tokens and ground-truth answer tokens. For the initialization of dataset-specific adapters during the commencement of continual learning (first dataset), we use the weights from FrozenBiLM, which is pre-trained on a substantial dataset comprising 10 million video-text pairs (WebVid10M~\citep{bain2021frozen}). In the training of domain-specific adapters for each subsequent domain, we conduct 20 epochs of training with an initial learning rate of $5e-5$. The learning rate undergoes a linear warm-up for the first 2 epochs, followed by a linear decay to 0.
Our proposed \Modelname~introduces only two hyperparameters. Specifically, we set the temperature parameter ($\tau$) to 0.01 and merge top-$k$=2 adapters for the adapter merging process. We normalize the router's predicted probabilities by setting the sum of the top-$k$ probabilities to 1 and the remaining probabilities to 0.

\textbf{Network Structures:} Our frozen pretrained backbone is FrozenBiLM~\citep{bain2021frozen}, comprising a language model DeBERTa-XL and a vision model CLIP-L/14. The input features to the router consist of the concatenation of the averaged hidden states from the 4th last layer of DeBERTa-XL and the averaged hidden states from the last layer of CLIP-L/14, without the incorporation of adapters. For each dataset, we employ an adapter comprising $L$ adapter layers, inserting an adapter layer after each self-attention layer and feed-forward network layer in DeBERTa-XL. Following~\citep{yang2022zero,houlsby2019parameter,yang2022zero}, each adapter layer in our approach includes a downsampling and an upsampling linear layer, along with a residual connection. The linear layers are configured with an 8$\times$ downsample scale to an intermediate hidden size, and the upsampler maps back to the original dimensionality.

\vspace{2mm}\noindent \textbf{Continual Learning Baselines.} Since our work is the very first exploration of continual VidQA learning, we implement a number of continual learning baselines (focused on image classification) to VidQA task, including three recent Prompt-based methods L2P~\citep{wang2022learning}, CODA-Prompt~\citep{smith2023coda}, and S-Prompts~\citep{wang2022s}and two regularization-based methods EwC~\citep{kirkpatrick2017overcoming} and LwF~\citep{li2017learning}. For a fair comparison, we use the same pretrained model and preserve most hyper-parameter settings with our approach. 

\begin{itemize}
    \item \noindent \textbf{L2P}~\citep{wang2022learning}. For the prompt settings, we set the prompt length to 10 and the size of the prompt pool to 6. The dimension of the prompt key is configured to be 3072, matching the dimension of the router input feature in our method. The prompt dimension is set to 1536, aligning with the input dimension of the frozen language model. We sweep the learning rate between $1e-2$ and $1e-5$ with an interval of 3.33$\times$. The best performance is achieved with an initial learning rate 3e-3.
    
    \item \noindent \textbf{CODA-Prompt}~\citep{smith2023coda}. For a fair comparison, we adopt the same prompt settings as our L2P baseline for CODA-Prompt. Following ~\citep{smith2023coda}, we apply orthogonality initialization to initialize the prompts, their keys, and their attention matrices. The dimension of prompt attention is set to 3072, consistent with the dimension of the prompt key. For optimal performance, we configure the learning rate to 1e-3.
    
    \item \noindent \textbf{S-Prompts}~\citep{wang2022s}. We use exactly the same prompt settings as in our implementation for L2P. For their K-Means router, we set $K=3$ as the number of centroids for each domain and 1-first-nearest neighbor with the centroids to search for the best prompts.
    
    \item \noindent \textbf{EwC}~\citep{kirkpatrick2017overcoming} and \textbf{LwF}~\citep{li2017learning}. We follow their original implementations, except that the regularization is only applied to adapters as all the other parameters are frozen. A single adapter is shared for all the domains.

\end{itemize}

\section{Evaluation Metrics}
\label{sup:eval_metrics}
Following \citep{wang2022s,wang2022dualprompt,wang2022learning,smith2023coda}, we employ standard evaluation metrics, including \textbf{average accuracy} and \textbf{forgetting}. The average accuracy metric evaluates both learning capacity and catastrophic forgetting, whereas the forgetting metric specifically addresses catastrophic forgetting. As an illustration, the pretrained zero-shot model attains 0\% forgetting but may exhibit relatively low average accuracy.

Formally, let $S_{t,\tau}$ denote the accuracy on the $\tau$-th dataset after training on the $t$-th dataset (task). After the training on the $t$-th dataset, we compute the average accuracy $A_t$ and forgetting $F_t$ as follows:

\begin{align}
A_t &= \frac{1}{t}\sum_{\tau=1}^{t}S_{t,\tau} \\
F_t &= \frac{1}{t}\sum_{\tau=1}^{t} \underset{\tau' \in \{1,...,t\}}{\max}(S_{\tau',\tau} - S_{t,\tau})
\end{align}
Upon completion of training on all $T$ datasets, we report the final average accuracy $A_T$ and forgetting $F_T$.

\section{Extension to Other Types of Continual Learning}
\label{sec:exp:other type cl}

To show the flexibility of our framework, we also extend \Modelname~to two other types of continual learning: 1) Class-Incremental Learning (CIL) and 2) Task-Incremental Learning (TIL) on VidQA.

\vspace{2mm}\noindent \textbf{CIL.} In standard CIL, there are no overlapping classes between tasks, and the training of each split or dataset is treated as a separate task. To simulate CIL, we treat each unique answer as a class, similar to the protocol commonly used in continual learning for image classification~\citep{wang2022learning}. We conduct experiments in two distinct settings: 1) \textit{MSRVTT-QA 10 subsets }, where the classes do not overlap between subsets, and 2) \textit{4-Datasets} (iVQA, MSVD, LSMDC, ActivityNet), excluding samples with answers that overlap across datasets. In the first setting, the model is continually trained on these disjoint subsets of the data, while in the second setting, we train our model on the 4 continually arriving datasets. The results, presented in Tab.~\ref{tab:cil}, show that \Modelname~consistently outperforms S-Prompts~\citep{wang2022s}, achieving \textbf{18.2\%} and \textbf{8.5\%} improvement on average accuracy on MSRVTT-QA 10-tasks and 4-Datasets respectively.

\vspace{2mm}\noindent\textbf{TIL.} To extend our framework to TIL, we treat the training on each dataset as a task. Unlike DIL or CIL, in TIL, each test sample during inference is provided with a dataset identity. As shown in Tab.~\ref{tab:til}, \Modelname~obtains only 0.1\% lower accuracy than the upper-bound multitask learning baseline. This is because in this setting, \Modelname~can always use the correct dataset-specific adapters, which are individually trained on their corresponding datasets and perform comparable to multitask training.

\begin{table}
    \centering
    \caption{Class-Incremental Learning (CIL) experiments are conducted under two settings: continually training our model on 1) 10 data subsets of MSR-VTT without overlapping classes (answers), and 2) 4 sequentially arriving datasets (iVQA, MSVD, LSMDC, ActivityNet) that do not have any overlap between their classes (answers). The proposed \Modelname~outperforms S-Prompts by a large margin in both settings.}
    \label{tab:cil}
    \begin{tabular}{lcccc}
    \toprule
    & \multicolumn{2}{c}{MSRVTT-QA 10 subsets} & \multicolumn{2}{c}{4-Datasets} \\
    Method	& Average Acc.&	Forgetting&	Average Acc.&	Forgetting\\
    \midrule
    Multitask (upper-bound)& 47.3&	-&	51.6&	-\\
    \midrule
    S-Prompts &	15.4	&-23.5&42.2	&-3.3\\
    \Modelname~(Ours)&	\textbf{33.6}&	\textbf{-13.7}	& \textbf{50.7}	& \textbf{-0.9} \\
    \bottomrule
    \end{tabular}
\end{table}

\begin{table}
    \centering
    \setlength{\tabcolsep}{4pt}
    \caption{Application of our model to Task-Incremental Learning (TIL). Our proposed framework generalizes well to TIL achieving only $0.1\%$ lower accuracy than the upper-bound multitask learning baseline.}
    \label{tab:til}
    \begin{tabular}{lcccccccc}
    \toprule
    Method	&iVQA&	MSVD&	MSR-VTT&	LSMDC	&ActivityNet	&TGIF&	Avg.\\
    \midrule
    Multitask (upper-bound)&	39.7&	\textbf{56.6}&	46.7&	62.9&	42.2&	67.8&	\textbf{52.6}\\
    \midrule
    \Modelname	& \textbf{39.8}&	54.8&	\textbf{46.7}	&\textbf{63.0}&	\textbf{42.4}	&\textbf{68.0}&	52.5\\
    \bottomrule
    \end{tabular}
\end{table}

\section{Dataset Descriptions}
\noindent \textbf{Video Question Answering(VidQA).} We perform experiments on 9 Video Question Answering (VidQA) datasets, which include iVQA~\citep{yang2021just}, MSVD-QA~\citep{xu2017video}, MSRVTT-QA~\citep{xu2017video}, LSMDC~\citep{maharaj2017dataset}, ActivityNet-QA~\citep{yu2019activitynet}, TGIF-QA~\citep{jang2017tgif}, TrafficQA~\citep{xu2021sutd}, EnvQA~\citep{gao2021env} and AGQA~\citep{grunde2021agqa}. MSVD-QA, MSRVTT-QA, and ActivityNet-QA involve social media videos, with ActivityNet-QA featuring notably longer videos (i.e., on average 2 minutes in length versus 30 second average duration of the videos in the first two datasets). iVQA, LSMDC, TGIF-QA, TrafficQA, EnvQA, and AGQA represent instructional, movie, short-GIF, traffic, virtual, and indoor human videos, respectively.

\begin{itemize}
    \item \noindent \textbf{iVQA}~\citep{yang2021just} is an open-ended VidQA dataset with reduced language biases and high-quality redundant manual annotations. It contains 10K video clips and 10K questions, split into 6K/2K/2K for training/validation/testing.
    \item \noindent \textbf{MSVD-QA}~\citep{xu2017video} is an open-ended VidQA dataset based on Microsoft Research Video Description Corpus~\citep{chen2011collecting}. It contains 1.8K video clips and 51K question-answer pairs, split into 32K/6K/13K for training/validation/testing.
    \item \noindent \textbf{MSRVTT-QA}~\citep{xu2017video} is an open-ended VidQA dataset based on MSR-VTT dataset~\citep{xu2016msr}. It contains 10K video clips and 243K question-answer pairs, split into 158K/12K/73K for training/validation/testing.
    \item \noindent \textbf{ActivityNet-QA}~\citep{yu2019activitynet} is an open-ended VidQA dataset based on long videos \citep{caba2015activitynet} (averaging 180 seconds) and human annotation. It contains 5.8K video clips and 58K question-answer pairs, split into 32K/18K/8K for training/ validation/ testing.
    \item \noindent \textbf{TGIF-QA}~\citep{jang2017tgif} is an open-ended VidQA dataset based on the Tumblr GIF (TGIF) dataset \citep{li2016tgif}. It contains 46K GIFs and 53K question-answer pairs, split into 39K/13K for training/testing.
    \item \noindent \textbf{LSMDC-FiB}~\citep{maharaj2017dataset} is an open-ended video-conditioned fill-in-the-blank task that consists of predicting masked words in sentences that describe short movie clips~\citep{rohrbach2015dataset}. It contains 119K video clips and 349K question-answer pairs, split into 297K/22K/30K for training/validation/testing.
    \item \noindent \textbf{Traffic-QA} \citep{xu2021sutd} is a dataset designed for video QA, comprising 10,080 in-the-wild videos and annotated with 62,535 QA pairs. It serves as a benchmark for assessing the cognitive capability of causal inference and event understanding models in complex traffic scenarios. Our experiments focus on \textit{setting-1/2}, where the model receives a question-answer pair as input and is tasked with predicting the correctness of the answer (yes or no).
    \item \noindent \textbf{Env-QA} \citep{gao2021env} is a new video QA dataset to evaluate the ability of understanding the composition, layout, and state changes of the environment presented by the events in videos. It contains 23.3K videos collected in AI2-THOR simulator and 85.1K questions.
    \item \noindent \textbf{AGQA} \citep{grunde2021agqa} is a benchmark for compositional spatio-temporal reasoning. AGQA contains 192M unbalanced question answer pairs for 9.6K videos. We experiment on AGQA-v2 that contains a balanced subset of 2.27M question answer pairs to mitigate language bias.
\end{itemize}

\noindent \textbf{Visual Question Answering(VQA).}  Follow~\citep{li2023blip2}, we evaluate our model on 4 mainstream VQA datasets: OK-VQA~\citep{marino2019ok}, aOK-VQA~\citep{schwenk2022okvqa}, GQA~\citep{hudson2019gqa} and VQAv2~\citep{goyal2017making}.

\begin{itemize}
    \item \noindent \textbf{OK-VQA}~\citep{marino2019ok} is a knowledge-based visual question-answering benchmark with 14k images and 14k questions. 
    
    \item \noindent \textbf{aOK-VQA}~\citep{schwenk2022okvqa} is an augmented successor of OK-VQA~\citep{marino2019ok} and contains a diverse set of 25K questions requiring a broad base of commonsense and world knowledge to answer. 
    
    \item \noindent \textbf{GQA}~\citep{hudson2019gqa} is a large-scale visual question-answering dataset with real images from the Visual Genome~\citep{krishna2017visual} dataset and balanced question-answer pairs. 
    
    \item \noindent \textbf{VQAv2}~\citep{goyal2017making} consists of 1.1M questions about COCO images~\citep{chen2015microsoft} each with 10 answers. It is the balanced version of the original VQA~\citep{antol2015vqa} dataset. 

\end{itemize}

\bibliography{iclr2023_conference}
\bibliographystyle{iclr2023_conference}

\end{document}